\documentclass[wcp]{jmlr}

\usepackage{longtable}
\usepackage{booktabs}
\usepackage{lineno}
\usepackage{algorithm}
\usepackage{algpseudocode}

\usepackage{amsmath}
\usepackage{graphicx}
\usepackage{float}

\makeatletter
\let\Ginclude@graphics\@org@Ginclude@graphics 
\makeatother

\jmlrvolume{}
\jmlryear{2024}

\title[Emergence in Neural Networks]{Quantifying Emergence in Neural Networks: Insights from Pruning and Training Dynamics}

\author{\Name{Faisal AlShinaifi} \Email{Falshinaifi@ucsd.edu}\\
\addr University of California, San Diego
\AND
\Name{Zeyad Almoaigel} \Email{zalmoaigel@ucsd.edu}\\
\addr University of California, San Diego
\AND
\Name{Johnny Jingze Li} \Email{jil164@ucsd.edu}\\
\addr University of California, San Diego
\AND
\Name{Abdulla Kuleib} \Email{akuleib@ucsd.edu}\\
\addr University of California, San Diego
\AND
\Name{Gabriel A. Silva} \Email{gsilva@ucsd.edu}\\
\addr University of California, San Diego
}

\begin{document}

\maketitle

\begin{abstract}
Emergence, where complex behaviors develop from the interactions of simpler components within a network, plays a crucial role in enhancing neural network capabilities. We introduce a quantitative framework to measure emergence during the training process and examine its impact on network performance, particularly in relation to pruning and training dynamics. Our hypothesis posits that the degree of emergence—defined by the connectivity between active and inactive nodes—can predict the development of emergent behaviors in the network. Through experiments with feedforward and convolutional architectures on benchmark datasets, we demonstrate that higher emergence correlates with improved trainability and performance. We further explore the relationship between network complexity and the loss landscape, suggesting that higher emergence indicates a greater concentration of local minima and a more rugged loss landscape. Pruning, which reduces network complexity by removing redundant nodes and connections, is shown to enhance training efficiency and convergence speed, though it may lead to a reduction in final accuracy. These findings provide new insights into the interplay between emergence, complexity, and performance in neural networks, offering valuable implications for the design and optimization of more efficient architectures.
\end{abstract}

\section{Introduction}

Eemergence is a phenomenon where complex behaviors arise from the interactions of simpler elements within the network. Understanding and leveraging emergence is critical for further enhancing the capabilities of artificial neural networks. However, there has been a notable lack of work on defining emergence and its practical applications for improving network architectures \cite{categorical_framework_emergent_effects}. Emergence in neural networks can be observed when intricate patterns and functionalities develop from the collective dynamics of individual neurons and layers \cite{Goodfellow2016}. This emergent behavior is not explicitly programmed but results from the training process, leading to the network's ability to solve complex tasks \cite{Nair2010}. In this paper, we delve into the theoretical aspects of emergence in neural networks and validate our findings through empirical studies.

Our primary objective is to quantify emergence during the training of neural networks and to investigate its impact on network performance. We focus on the connectivity between active and inactive nodes within the network, hypothesizing that the quantity of this emergence is a predictor of the development of emergent traits \cite{Han2015}. To validate this hypothesis, we conduct extensive experiments on Multi-Layer Perceptrons (MLPs) and Convolutional Neural Networks (CNNs) using benchmark datasets such as MNIST,  Fashion-MNIST, and CIFAR-10 \cite{Glorot2011}.

The results of our experiments indicate a strong correlation between higher levels of emergence and improved trainability and performance of the networks \cite{Ruder2016}. Networks exhibiting greater emergence tend to converge more efficiently and achieve higher accuracy. Additionally, we explore the concept of network complexity and its spatial representation within the loss landscape. Emergence is found to reflect the concentration of potential local minima, suggesting that networks with higher emergence can navigate the loss landscape more effectively \cite{Goodfellow2016}.

Pruning, a technique used to reduce network complexity by eliminating non-essential nodes and connections, is also examined in the context of emergence \cite{Han2015}. While pruning leads to faster convergence and enhanced training efficiency, it typically results in a reduction of final accuracy. This trade-off highlights the importance of balancing complexity and efficiency in the design of neural network architectures \cite{Nair2010}.


 \textbf{Emergence Increases with Complexity:} Emergence (\(E\)) is inherently tied to the network’s overall complexity, which increases with the number of parameters and layers. Higher emergence values in larger, more complex networks indicate a greater potential for developing sophisticated behaviors and patterns during training. Intuitively, complexity in this context refers to the network's capacity to represent intricate functions and patterns, which is often a function of its architecture and the number of parameters it possesses. Emergence, as a measure of complexity, suggests that as networks become more complex, they have a higher potential for exhibiting emergent traits, leading to more nuanced and sophisticated behaviors \cite{categorical_framework_emergent_effects}.

We have also defined a relative emergence, which investigates the emergence of the model relative to the number of parameters in the model, which gives us a way to find emergence relative to the size of network. 

 \textbf{Relative Emergence Correlates with Trainability:} Relative emergence (\(\tilde{E}\)) provides a metric for evaluating how efficiently a network’s complexity contributes to its learning process. Trainability refers to the ease and efficiency with which a neural network can learn from data and improve its performance during training. A higher relative emergence in pruned networks suggests that these networks, although simpler, are more adept at learning and adapting, leading to faster convergence and improved performance during training. This can be understood as pruned networks, despite having fewer parameters, are better at leveraging their complexity for effective learning, thus being more trainable \cite{Ruder2016}.


Additionally, we discuss the interpretation of emergence from the loss function landscape perspective. Models with strong emergence can be trained on a range of different tasks, which suggests the existence of multiple local minima. If emergence value is low, it could indicate a flat region of the loss function landscape, where high emergence value suggests a more rugged, complex region. This insight allows us to understand the local geometry of the loss landscape and predict the training behavior of the network.

Our work builds on and is inspired by recent theoretical advancements, such as the framework presented by \cite{categorical_framework_emergent_effects}. This work provides a rigorous mathematical foundation for understanding emergence in network structures, which we extend and apply to the training dynamics of neural networks. By integrating these theoretical insights with empirical validation, our findings offer new perspectives on the role of emergence in neural network performance and complexity. Our findings offer new insights into the theoretical underpinnings of emergence and complexity in neural networks. By understanding how emergence influences network performance and complexity, we can develop more efficient and effective neural network architectures.

\section{Related work}

The concept of emergence in neural networks has been explored extensively in various contexts. \citet{emergence_hierarchical_structure} classify emergence as the development of hierarchical modularity within complex systems, where smaller, function-specific modules combine to form larger, complex structures. This hierarchical structure is seen as a product of evolutionary processes that optimize the system for efficiency and robustness, often resulting in an "hourglass architecture" where the system produces many outputs from many inputs through a relatively small number of highly central intermediate modules.

Emergence is also framed similarly by \citet{hierarchical_route_emergence_leader_nodes}, who define emergence within the context of network structures, particularly focusing on the emergence of leader nodes in non-normal networks. In this context, emergence is tied to the directedness and asymmetry of the network, where as the network becomes more non-normal, leader nodes (nodes with no out-degree) spontaneously emerge, driving the dynamics and hierarchical organization within the network. Another perspective on emergence focuses on emergent abilities through the lens of pre-training loss rather than model size or training compute, as discussed by \citet{emergent_abilities_loss_perspective}.

Emergence can also be defined in a mathematical sense, particularly concerning networks. \citet{categorical_framework_emergent_effects} provide a categorical framework for quantifying emergent effects in complex systems through the lens of network topology. Their framework introduces a computational measure of emergence that ties the phenomenon to the network's topology and local structures. This approach offers a novel method for quantifying emergence, which could be applied to a wide range of systems, including machine learning models and biological networks.

Another phenomenon closely related to emergence is the robustness of neural networks. \citet{training_data_accuracy_robustness} examine the relationship between the size of training data and the resulting accuracy and robustness of neural networks. Robustness refers to the ability of a neural network to maintain its performance in the presence of small, imperceptible changes to the input data, which can cause a well-trained model to make incorrect predictions. It is hypothesized that a robust neural network should exhibit no emergent traits. As networks become more specialized through training, they may lose some generalization capability (or overall robustness), similar to the decline or stabilization of robustness.

Additionally, \citet{robustness_neural_networks_probabilistic_approach} explore probabilistic robustness, which accounts for real-world input distributions and provides a practical approach for verification. Robustness ensures that the network performs reliably, and studying emergence explores the internal dynamics that contribute to this performance.

\section{Methodology}

\subsection{Emergence as a Predictor}

Emergence has been widely studied in neural networks as the new property/ abilities of the network as the size of the network grows. The natural question here arises, as predicting what scale would allow the network to exhibit these emergent properties. To answer this question, it is important to give an approximation of the size of the network sufficient to train on a dataset. In this section, we built a measure of emergence, which is found to be correlating with the model's ability to train on a given dataset. 
Emergence is hypothesized to predict the network's future training trajectory by evaluating changes in node activations and weights. By observing these changes during the early stages of training, we can estimate the network's potential to develop complex, emergent traits. Specifically, we calculate emergence measures based on the connectivity between active and inactive nodes, which serve as indicators of the network’s potential for emergent behaviors. The dichotomy classification of the active and inactive nodes is motivated by exploring the feature formation inside the network, where nodes with higher activations are representing certain features \cite{simonyan2013deep, bau2017network, yosinski2015understanding}.

\subsection{An Overview of the Mathematical Framework for Emergence}

Emergence fundamentally arises from the observation of a system from a higher scale. According to \citet{adam2017systems}, emergent effects are defined as:

\[
\Phi(s_1 \lor s_2) \neq \Phi(s_1) \lor \Phi(s_2)
\]

for some constituent subsystems \( s_1 \) and \( s_2 \). Here, \( s_1 \) and \( s_2 \) are components of the system, \( \Phi \) represents the observation or computational process, and \( \lor \) is the binary operation encoding the interactions among the components. In machine learning, \( s_1 \) and \( s_2 \) can be models or datasets, and \( s_1 \lor s_2 \) represents the combined model/dataset. Emergent phenomena occur when the effect of the combined model/dataset differs from the sum of the separate models/datasets.


For our study, we focus on a single model's potential to exhibit emergent effects. \citet{categorical_framework_emergent_effects} proposed a network-based measure of emergence as:

\[
\text{Emergence}(G,H) = \sum_{x \in G \setminus H} \#\text{paths in } H \text{ from } N_H(x) \text{ to } H
\]

where:
$ G$ is the network at the lower scale,
$ H$ is the image of $G$ under the mapping $\Phi$, representing the representation of the network at the higher scale.


In order to study emergence during the training of neural networks, we represent the initial network as \( G \) and the trained network as \( H \). If we use \( H \) to represent the subnetwork whose nodes remain active in the trained neural network, 

In machine learning setting, one modeling approach is to consider $\Phi$ as the training process, since emergence here evaluates the potential/ ability for emergent traits when we observe system $G$ from a higher level $H$, here we want $G$ to represent the model itself, and $H$ to be some certain features of the model. In the paper, we adopt the setting that $H$ is the nodes in $G$ that are active in the training process, where active nodes are defined as the set of nodes whose activation is greater than a threshold set close to $0$. This sorted out the nodes that are not actively participating in the computational process.  The set of active nodes thus in a sense represent the learning task, thus we can tie emergence with the performance of the network in a learning process. This fits in our framework of emergence, where part of the system is being neglected after the learning process, thus the learning process represents the $\Phi$ where partial observation is carried out, and the properties of $H$ represents the emergent abilities of the network.

we have the following measure of emergence for neural networks:

\[
E = \sum_{i=1}^{N-1} \sum_{j>i}^{N} (n_i - a_i) a_j \prod_{k=i+1}^{j-1} n_k
\]

where \( N \) is the number of layers, \( n_i \) is the total number of nodes in layer \( i \), \( a_i \) is the number of active nodes in layer \( i \).

For convolutional networks, the information flow is constrained by the pooling layers, so we have:

\[
E = \sum_{i=1}^{N-1} \sum_{j>i}^{N} (n_i - a_i) a_j \prod_{k=i+1}^{j-1} m_k
\]

where \( m_k \) is the number of filters in layer \( k \).


Note that here, emergence is only a function of the number of active nodes because of the good symmetry in MLP or convolutional networks. For other types of architectures, the equation to compute emergence will need to be modified based on their specific structures.

When the network architecture is fixed, meaning \( L \) and \( n_i \) for \( i = 1, \ldots, L \) and \( m_k \) for all pooling layers are fixed, emergence is only a function of the number of active nodes in each layer:

\[
E := E(a_1, \ldots, a_N)
\]


The number of active nodes at initialization is impacted by the weights. With a criterion for active nodes, for example, those nodes whose activation is greater than a threshold as adopted in this paper, we can establish an activation-based measure of emergence in neural networks.

\subsection{Measuring Emergence in Neural Networks}

Emergence in neural networks is quantified by counting the number of paths between 'alive' and 'dead' nodes. This measure reflects the network's complexity and its capacity to develop emergent traits. By analyzing the emergence activation over time, we can infer the network's potential for learning and adapting to new patterns.

We define emergence \( E \) as a function of the number of paths between active (alive) and inactive (dead) nodes, where a node is considered alive if its activation exceeds a certain threshold \( \theta \) and dead otherwise:

\[
E = f(\text{paths}_{\text{inactive-active}})
\]
\[
\text{Alive if } a^{(l)} > \theta, \text{ Dead if } a^{(l)} \leq \theta
\]

This measure of emergence reflects the complexity and potential for the network to exhibit emergent traits as training progresses. A higher emergence value indicates a higher likelihood of complex behavior.

Node activation is a fundamental concept in neural networks, representing the output of a node after applying an activation function. The activation function introduces non-linearity into the network, enabling it to learn complex patterns. Common activation functions include ReLU (Rectified Linear Unit), Sigmoid, and Tanh \cite{Nair2010}.

To detect the activation of a node, we monitor its output value after applying the activation function. During each epoch, we record the activation levels of all nodes across different layers. This data is then analyzed to determine the proportion of alive and dead nodes, providing insights into the network's learning dynamics and complexity \cite{Glorot2011}.

Throughout the training process, we observe a general trend of decreasing emergence activation, indicating that the network becomes more specialized and focused on relevant features. Pruning, which involves systematically removing nodes and connections that contribute minimally to the network’s performance, further reduces emergence activation. This reduction signifies a decrease in complexity, correlating with faster convergence and improved training efficiency \cite{Han2015}.

\subsection{Emergence, Relative Emergence and Training Dynamics}

Emergence in neural networks can be thought of as reflecting the complexity of the network, providing insights into how the future training process may unfold. Specifically, emergence serves as a predictor of the network's capabilities by analyzing the initial weights and activations during the early epochs of training. The fundamental idea is that emergence quantifies how changes in a small component of the network, such as individual nodes, influence the network's overall behavior. This is measured through changes in activation levels and weights.


The impact of emergence on the training dynamics and loss-function landscape can be understood as follows. In the previous sections, we quantify emergence as the number of paths from the inactive nodes to the active nodes; it essentially captures the potential paths of information flow that form feature representation in the network. If the network has higher value of emergence, there are more paths between base-nodes (inactive) to the feature representing nodes (active), the features are formed by summarizing over larger amount of inputs. When networks exhibit such characteristics, intuitively it is more likely to form features \citet{shwartz2022information}, which corresponds to more locally minimums \citet{li2018visualizing}. 

The concept of relative emergence provides a nuanced understanding of the interplay between network complexity and trainability. While absolute emergence is a measure of the overall complexity of a neural network, relative emergence offers a normalized metric that accounts for the network's size. This section elucidates how our theory correlates with relative emergence and its implications for network pruning and trainability.

In our study, we observe that pruned networks exhibit smaller absolute emergence, reflecting their reduced complexity due to the decreased number of parameters. However, when we normalize emergence by the number of parameters, defining this normalized metric as relative emergence, denoted \( \tilde{E} \), we gain deeper insights into the network's trainability:

\[
\tilde{E} = \frac{E}{\text{\# parameters}}
\]

This relative emergence, \( \tilde{E} \), tends to be larger in pruned networks compared to non-pruned ones. This increase in relative emergence indicates that, despite the reduced absolute complexity, pruned networks maintain a high degree of effective complexity per parameter, which correlates with stronger trainability. In other words, pruned networks, with fewer parameters, are more efficient in developing emergent traits that enhance their learning and generalization capabilities.

The implications of this relationship are twofold:
\begin{itemize}
    \item \textbf{Emergence Increases with Complexity}: Absolute emergence (\( E \)) is inherently tied to the network's overall complexity, which increases with the number of parameters and layers. Higher emergence values in larger, more complex networks indicate a greater potential for developing sophisticated behaviors and patterns during training.
    \item \textbf{Relative Emergence Correlates with Trainability}: Relative emergence (\( \tilde{E} \)) provides a metric for evaluating how efficiently a network's complexity contributes to its learning process. A higher relative emergence in pruned networks suggests that these networks, although simpler, are more adept at learning and adapting, leading to faster convergence and improved performance during training.
\end{itemize}

\begin{figure}[H]
\centering
\includegraphics[width=0.3\textwidth]{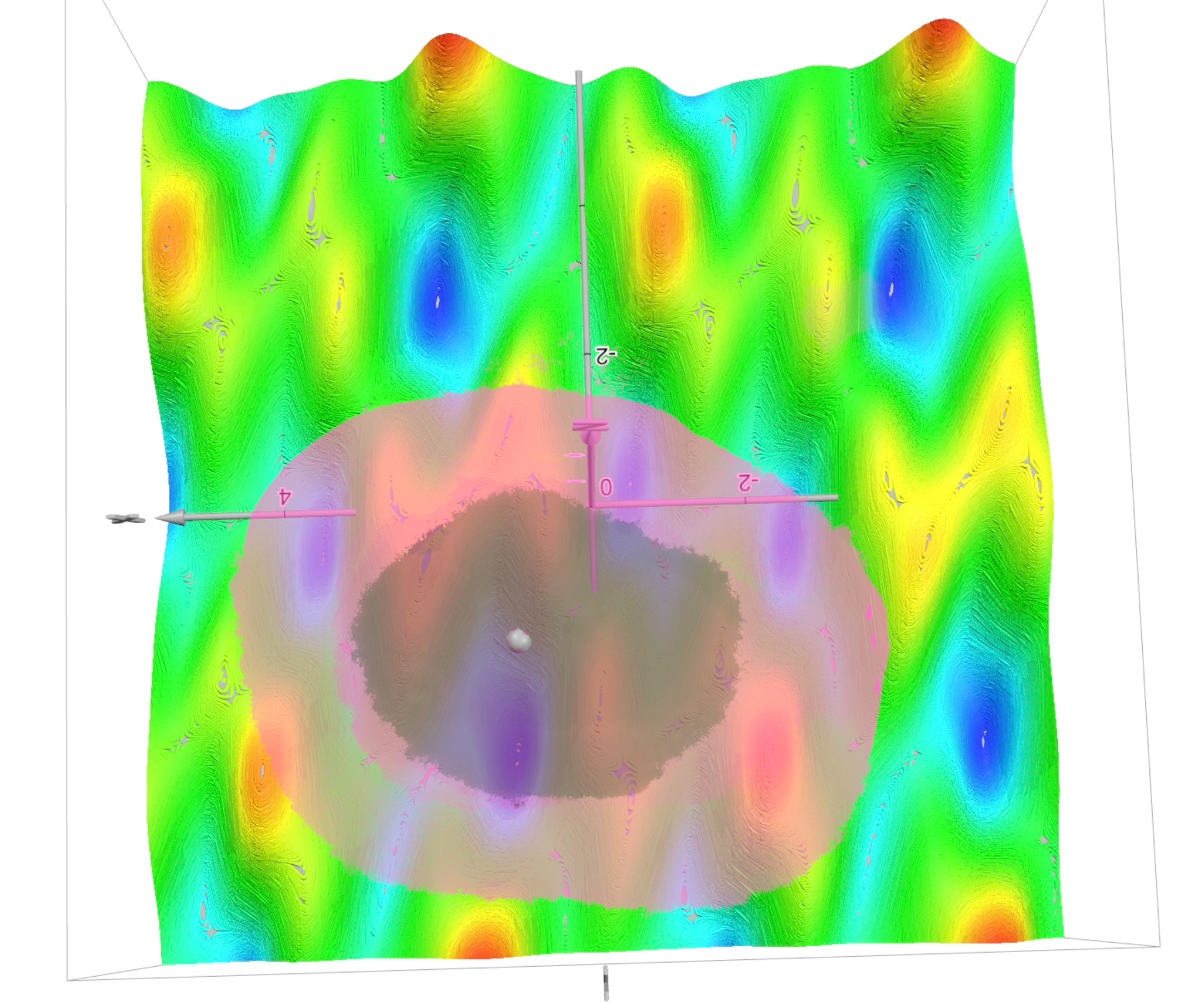}
\caption{Visualization of Emergence in the loss landscape}
\label{fig:enter-label}
\end{figure}

\subsubsection{Interpreting Emergence from the Energy Landscape Perspective}

Relative emergence can be conceptualized as reflecting the density of local minima within a given region of the loss landscape. Although pruning reduces the overall 'size' of the region by decreasing the number of parameters, it effectively increases the density of local minima within this smaller region. This higher density of local minima implies that the pruned network, despite its reduced complexity, possesses a more intricate and rich structure of optimization paths, facilitating faster and more efficient training convergence.

For \textit{\textbf{Figure 1}}, we explore how one would visualize emergence in the loss landscape pre- and post-pruning. The lighter pink region shows how the more complex pre-pruned network covers a larger region than the darker region. The blue region represents the local minima. This distinction is critical because it highlights that while absolute complexity (emergence) is essential for developing complex behaviors, the efficiency of this complexity (relative emergence) is crucial for practical trainability. By pruning the network, we reduce unnecessary parameters, thereby enhancing the network's ability to focus on the most relevant features and pathways, which accelerates learning and convergence.

In the context of our methodology, this understanding aligns with our findings on the spatial representation of network complexity within the loss landscape. Emergence reflects the size of the region around the network's current state, indicating potential local minima. Pruned networks, with their higher relative emergence, navigate this landscape more effectively, finding optimal solutions with greater efficiency.

Overall, the concept of relative emergence not only corroborates our theoretical framework but also provides actionable insights for optimizing neural network architectures, balancing complexity with trainability to achieve more efficient and effective learning models. We hypothesize that the quantity of emergence, represented by the number of paths between active and inactive nodes, is indicative of the potential for emergent traits to arise later in the training process. 


Furthermore, we propose that pruning the network will decrease the quantity of emergence due to its reduced complexity. This reduction in emergence is expected to correlate with faster convergence in training and improvements in training efficiency and accuracy.

\subsection{Emergence and Pruning and Its Impact}

Pruning is a technique used to enhance the efficiency of neural networks by reducing their complexity. In our experiments, pruning was implemented by systematically removing a significant portion of the weights and nodes that contribute minimally to the network’s performance. After training a model to achieve satisfactory accuracy, pruning is conducted. Post-pruning, we typically observe a drop in emergence, which reflects the drop in the dimension of the parameter space and a decrease in the network’s complexity. However, an effectively pruned network retains local minima within this reduced region, which accounts for the faster convergence observed. 

To determine the effectiveness of pruning, we utilize the concept of relative emergence. This involves assessing the density of local minima within the smaller region post-pruning. The expectation is that a well-pruned network will exhibit a higher ratio of local minima within its confined region compared to a non-pruned network. Consequently, the relative emergence of a pruned network should be higher, indicating a more complex loss landscape within the reduced region. These observations align with our broader theory and support the idea of spatial awareness in the emergence equation. By effectively pruning at the right moment, we enhance the network's specialization within a smaller radius. This facilitates faster convergence on similar datasets, underscoring the importance of timing and methodology in the pruning process to leverage these benefits fully. It is crucial to note that the specialization effect of pruning is only realized when pruning is conducted after an appropriate amount of training. This ensures that the network has sufficiently learned from the data before its structure is refined. Thus, effective pruning, combined with adequate training, leads to a network that is not only smaller but also more specialized and efficient within its operational region.

\section{Experiments, Results, and Discussion}

In our experiments, we conducted a comprehensive analysis of Multi-Layer Perceptrons (MLPs) to assess the impact of pruning on emergence and training dynamics. Initially, the models were trained for 5 epochs on both the MNIST and Fashion-MNIST datasets to establish a baseline performance, achieving initial accuracies of 90.4\% and 82\% respectively. Following this baseline, we created four identical copies of the trained model. These models were then subjected to different conditions: the first model continued training without pruning, serving as the control, with final accuracies of 95.7\% and 86.3\% respectively; the second, third, and fourth models were pruned by 30\%, 50\%, and 70\%. The pruning was executed using magnitude-based pruning to systematically reduce the network’s complexity. The learning rate across all models was consistently maintained at 0.005 to ensure uniform training conditions. Nodes were classified as 'alive' if their activation exceeded 0.05, with those below this threshold deemed 'dead'. Emergence was quantified by counting the number of paths between alive and dead nodes, which served as a proxy for the network’s complexity and its capacity for developing emergent traits.

\begin{figure}[H]
\centering
\begin{minipage}{0.48\textwidth}
\centering
\includegraphics[width=\textwidth]{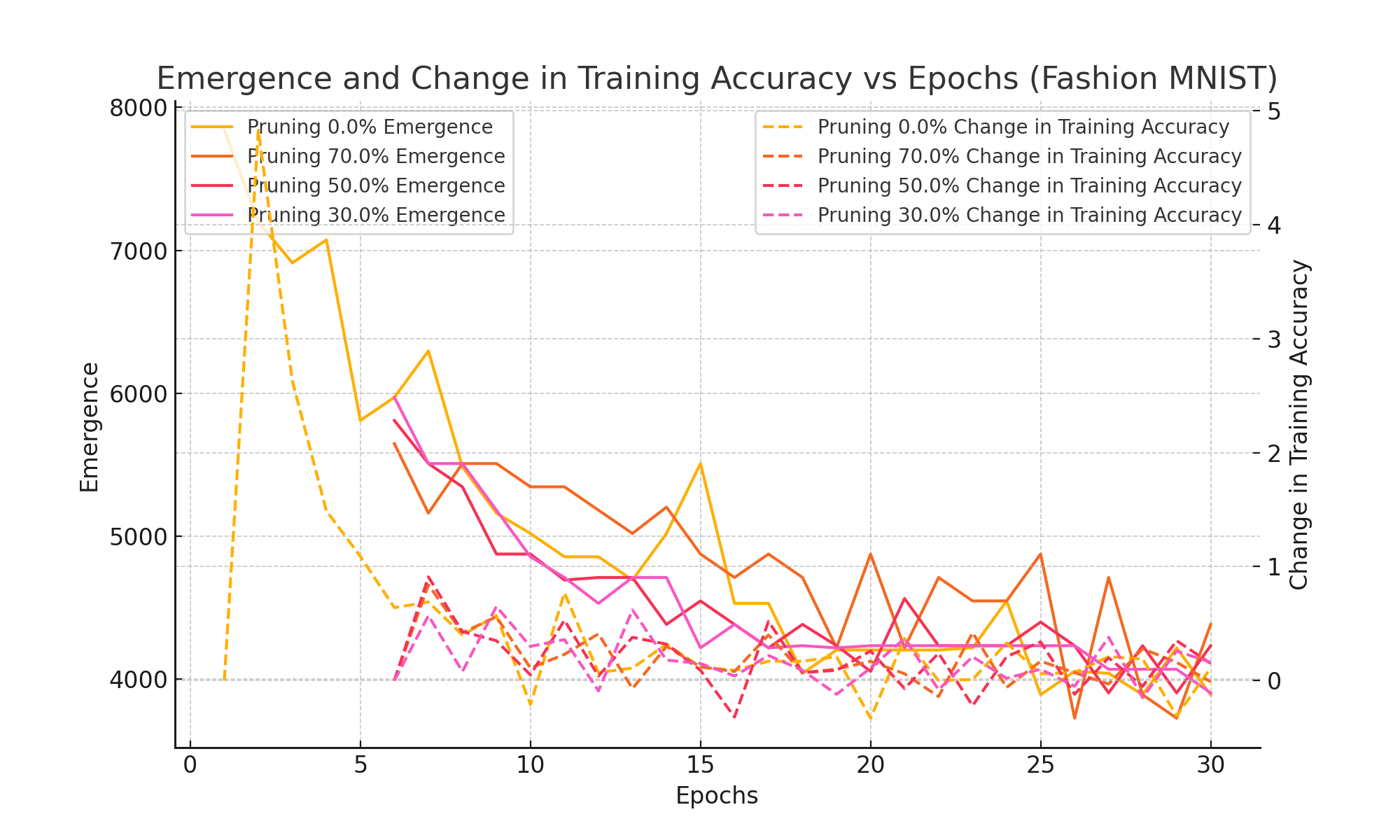}
\caption{Emergence and Change in Training Accuracy vs Epochs ( Fashion-MNIST)}
\end{minipage}
\hfill
\begin{minipage}{0.48\textwidth}
\centering
\includegraphics[width=\textwidth]{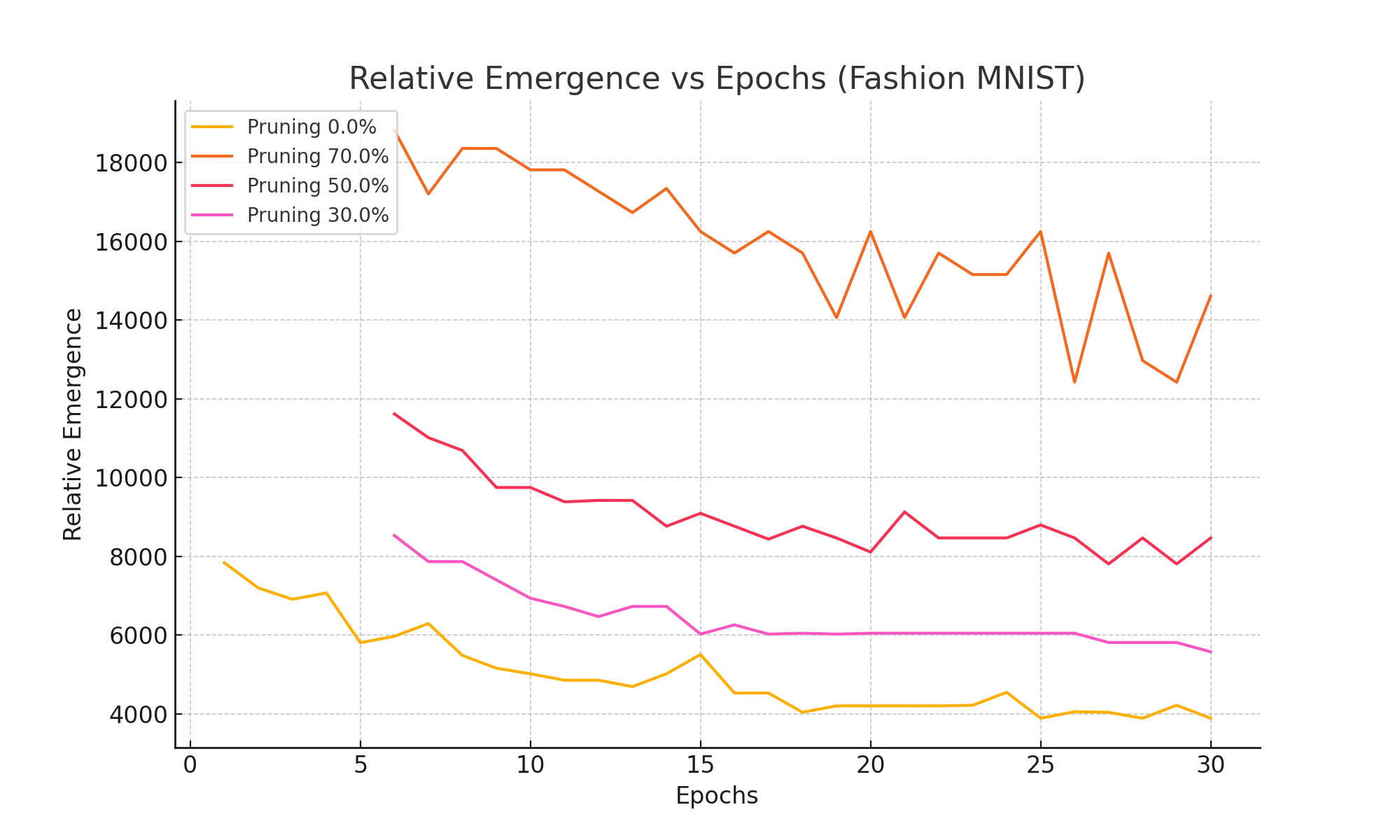}
\caption{Relative Emergence vs Epochs ( Fashion-MNIST)}
\end{minipage}
\end{figure}

\begin{figure}[H]
\centering
\begin{minipage}{0.48\textwidth}
\centering
\includegraphics[width=\textwidth]{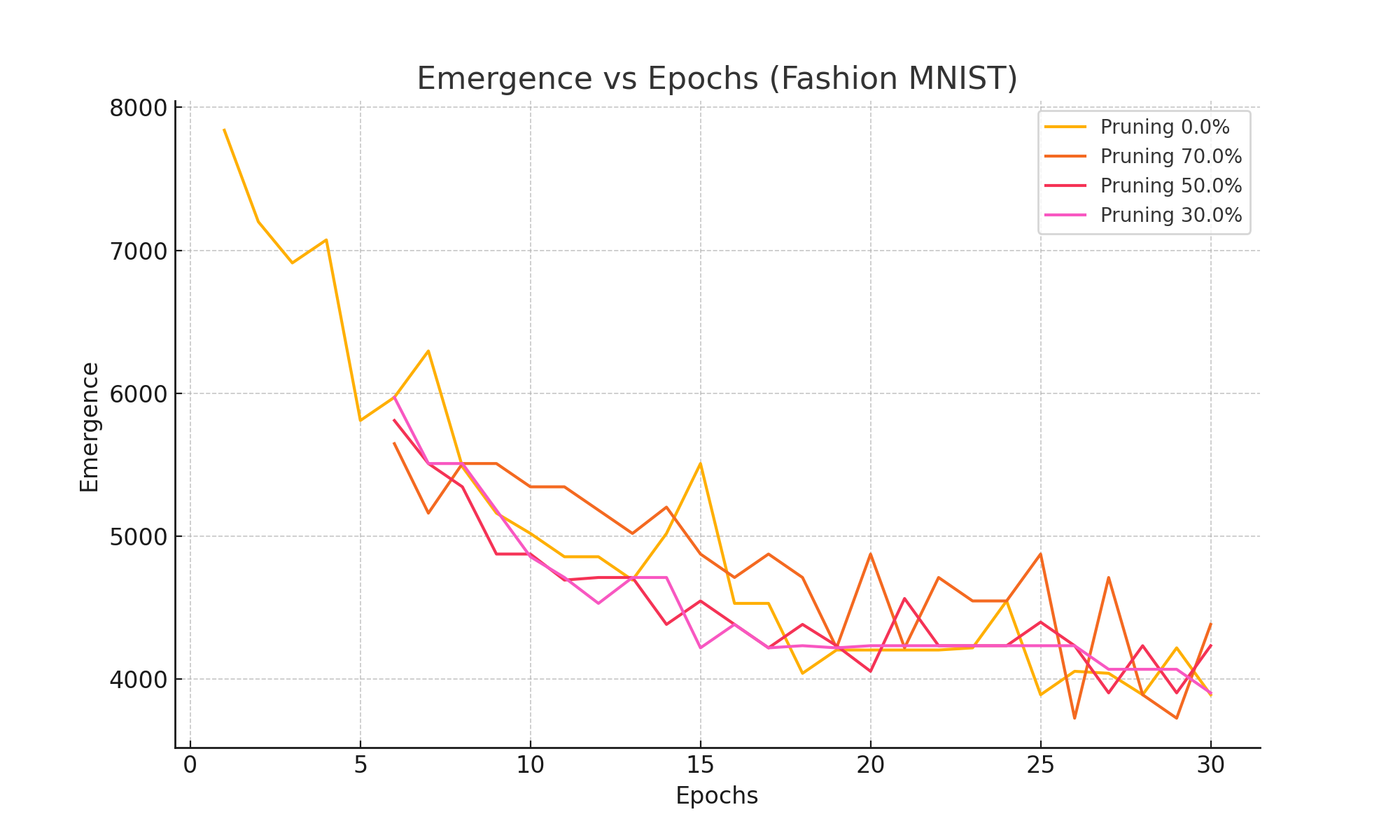}
\caption{Emergence vs Epochs ( Fashion-MNIST)}
\end{minipage}
\hfill
\begin{minipage}{0.48\textwidth}
\centering
\includegraphics[width=\textwidth]{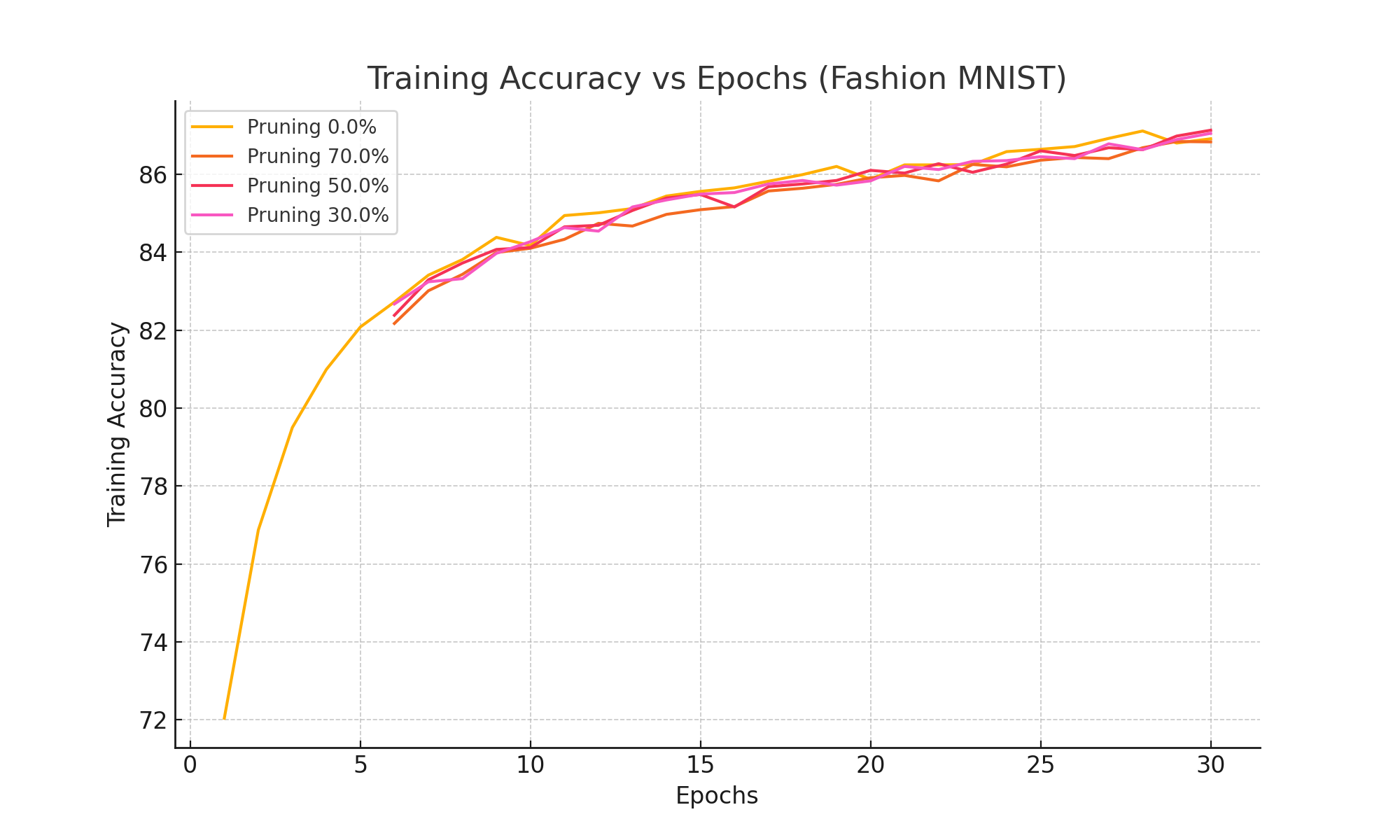}
\caption{Training Accuracy vs Epochs ( Fashion-MNIST)}
\end{minipage}
\end{figure}

\subsection{Fashion-MNIST Result Analysis}
The results obtained from the Fashion-MNIST dataset reveal a consistent trend: as emergence decreased, training accuracy increased, thereby supporting our hypothesis that emergence functions as a measure of potential within a neural network. Specifically, the control model, which underwent no pruning, reached a final accuracy of 86.3\%, while the 30\%, 50\%, and 70\% pruned models achieved final accuracies of 87\%, 86.8\%, and 86.2\% respectively. The decrease in emergence corresponded with increased network specialization, as evidenced by improved accuracy metrics. Pruning significantly reduced absolute emergence due to the smaller network size; however, relative emergence—normalized by the number of parameters—actually increased. This increase suggests that the pruned networks became more specialized, focusing on fewer but more relevant features. This enhanced specialization is reflected in the quicker convergence and improved training efficiency observed in the pruned models. 

Figures 2 to 5 illustrate these dynamics, showing the relationship between emergence, training accuracy, and the number of epochs. Notably, the 70\% pruned model, despite its significantly reduced complexity, exhibited a remarkably high relative emergence, indicating a more concentrated focus on essential learning pathways. This model achieved a faster convergence rate than its less pruned counterparts, albeit with a trade-off in final accuracy.

\begin{figure}[H]
\centering
\begin{minipage}{0.48\textwidth}
\centering
\includegraphics[width=\textwidth]{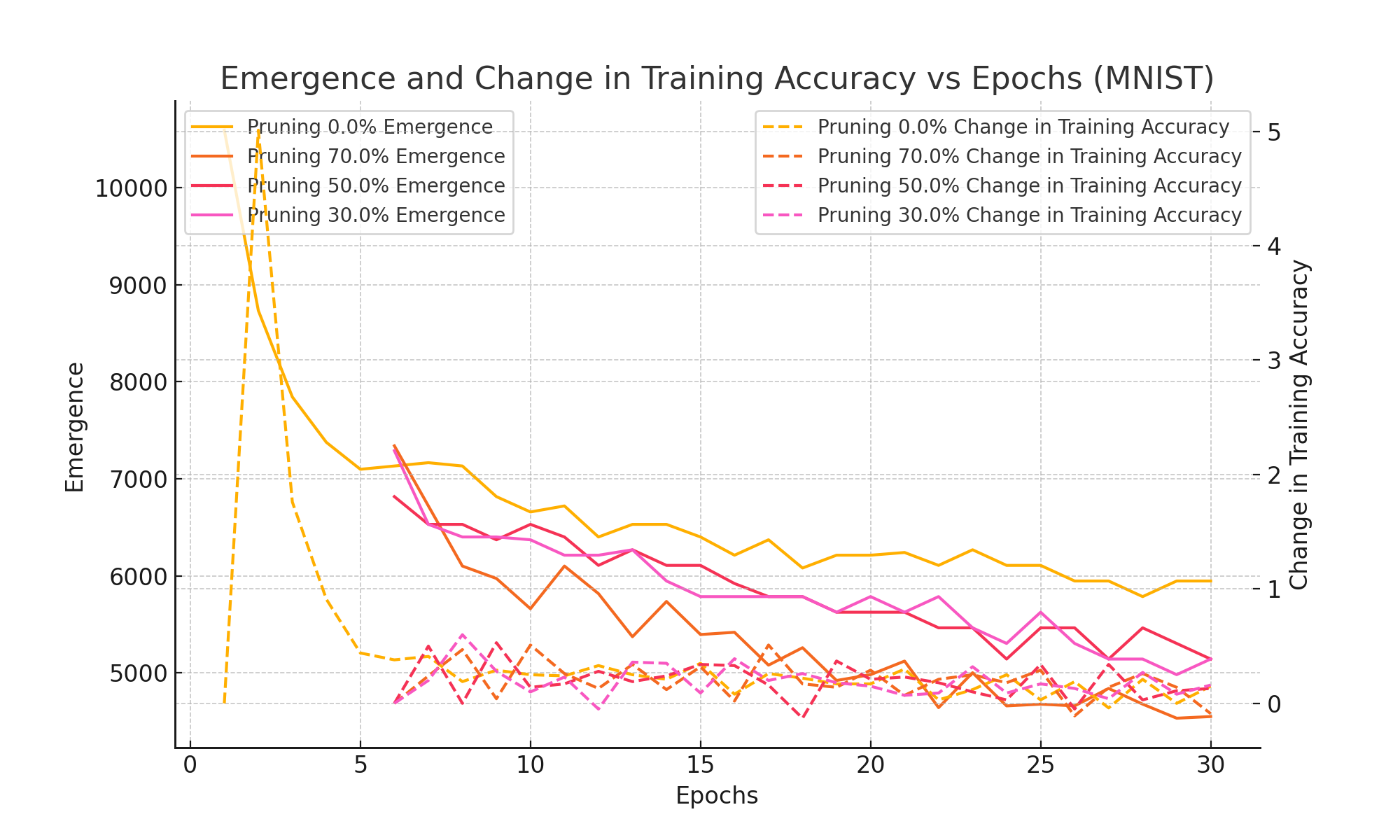}
\caption{Emergence and Change in Training Accuracy vs Epochs (MNIST)}
\end{minipage}
\hfill
\begin{minipage}{0.48\textwidth}
\centering
\includegraphics[width=\textwidth]{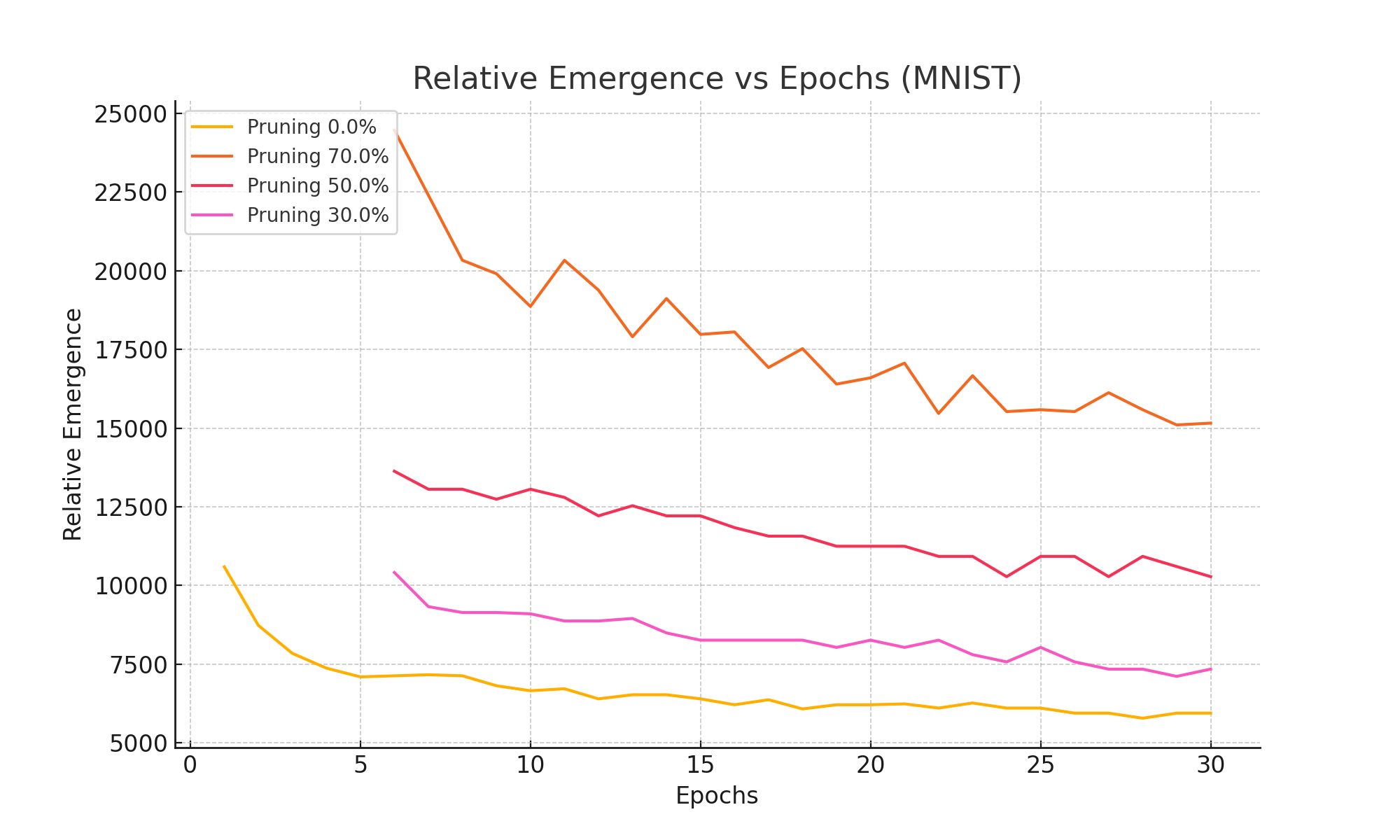}
\caption{Relative Emergence vs Epochs (MNIST)}
\end{minipage}
\end{figure}

\begin{figure}[H]
\centering
\begin{minipage}{0.48\textwidth}
\centering
\includegraphics[width=\textwidth]{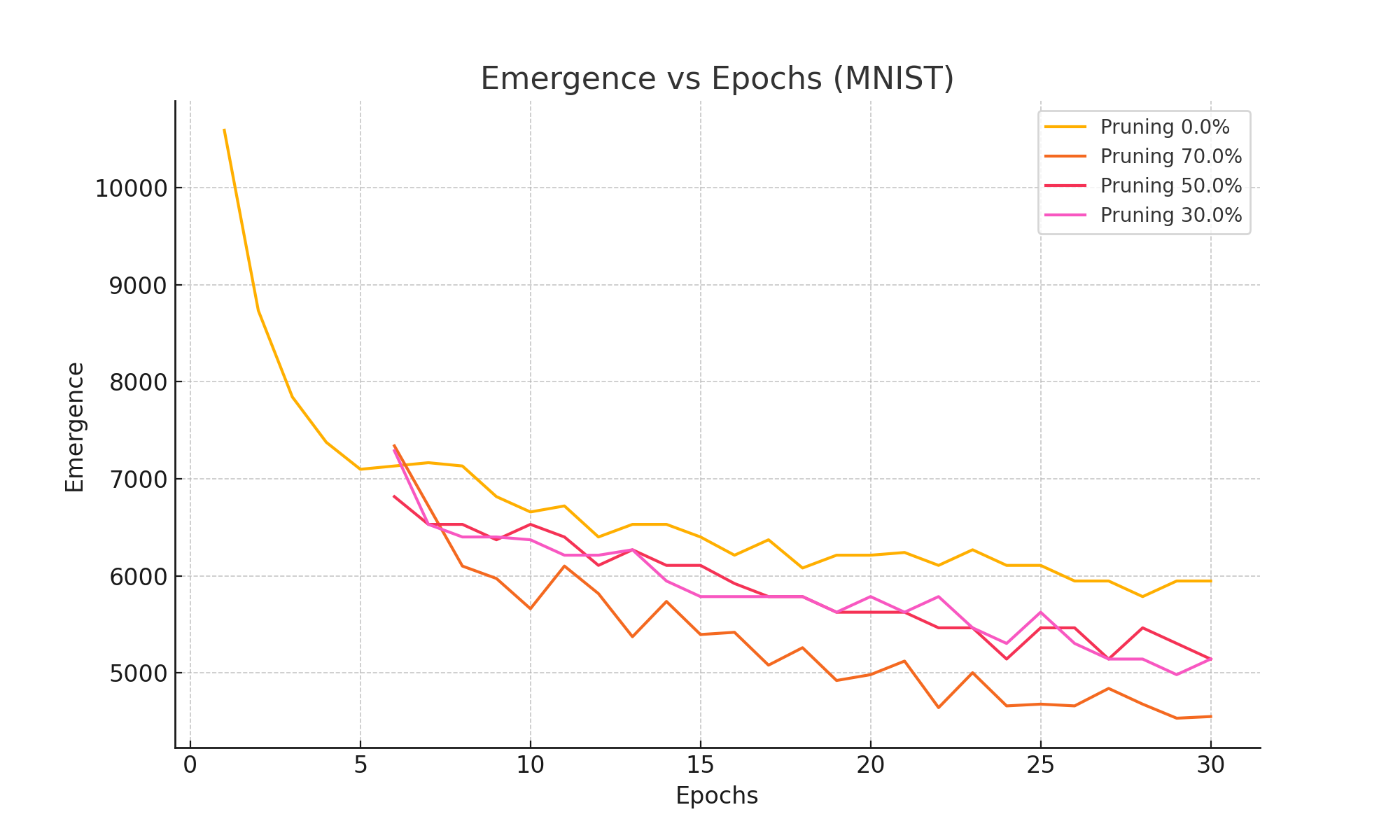}
\caption{Emergence vs Epochs (MNIST)}
\end{minipage}
\hfill
\begin{minipage}{0.48\textwidth}
\centering
\includegraphics[width=\textwidth]{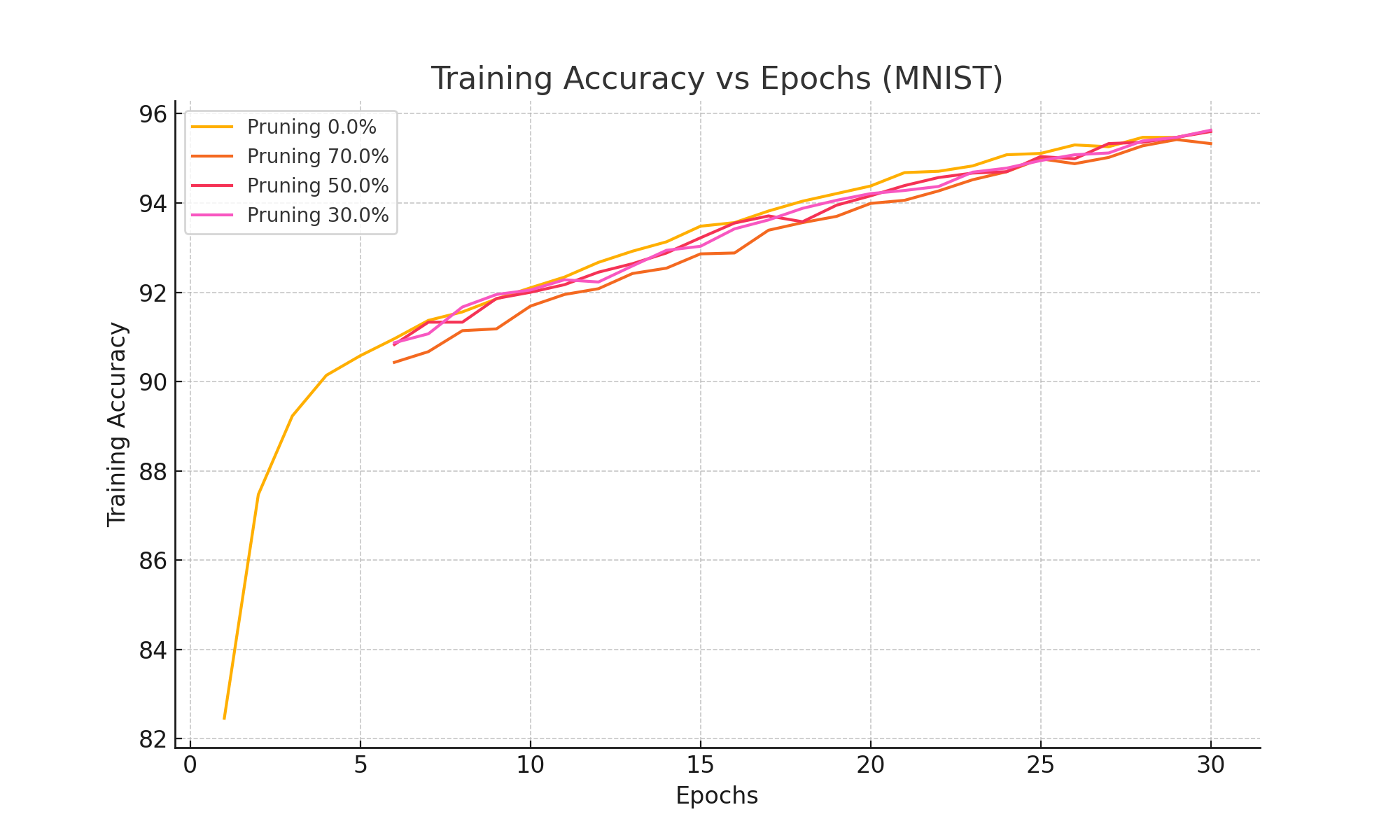}
\caption{Training Accuracy vs Epochs (MNIST)}
\end{minipage}
\end{figure}

\subsection{MNIST Result Analysis}
The MNIST dataset results exhibited the same trends observed in the Fashion-MNIST experiments, further validating our hypothesis. As emergence decreased, training accuracy improved, with the control model achieving a final accuracy of 95.7\%, and the pruned models achieving 95.7\%, 95.6\%, and 95.1\% respectively. The decrease in absolute emergence was consistent across all pruned models, yet relative emergence increased, particularly in the 50\% and 70\% pruned networks. This suggests that these pruned models, while simpler, had become more efficient in their learning processes, leveraging their remaining complexity more effectively. Figures 6 to 9 depict the same results as figures 2 to 5. 

\subsection{Impact of Pruning}
To further explore the impact of pruning on network performance and emergence, we conducted a series of experiments on both the MNIST and Fashion-MNIST datasets. The network was initially trained for 20 epochs and then split into two branches: one without pruning and one that was pruned. Both branches were then further trained for another 20 epochs. We also created two randomly initialized networks, one the size of the non-pruned network that was pre-trained on MNIST, and one that is the size of the pruned network. For the pre-trained models, the pruned network demonstrated faster convergence with lower emergence activation compared to its non-pruned counterpart. In contrast, the randomly initialized networks exhibited lower emergence and accuracy. This rapid convergence suggests that pruning accelerates the training process by reducing network complexity.

\begin{figure}[H]
\centering
\begin{minipage}{0.48\textwidth}
\centering
\includegraphics[width=\textwidth]{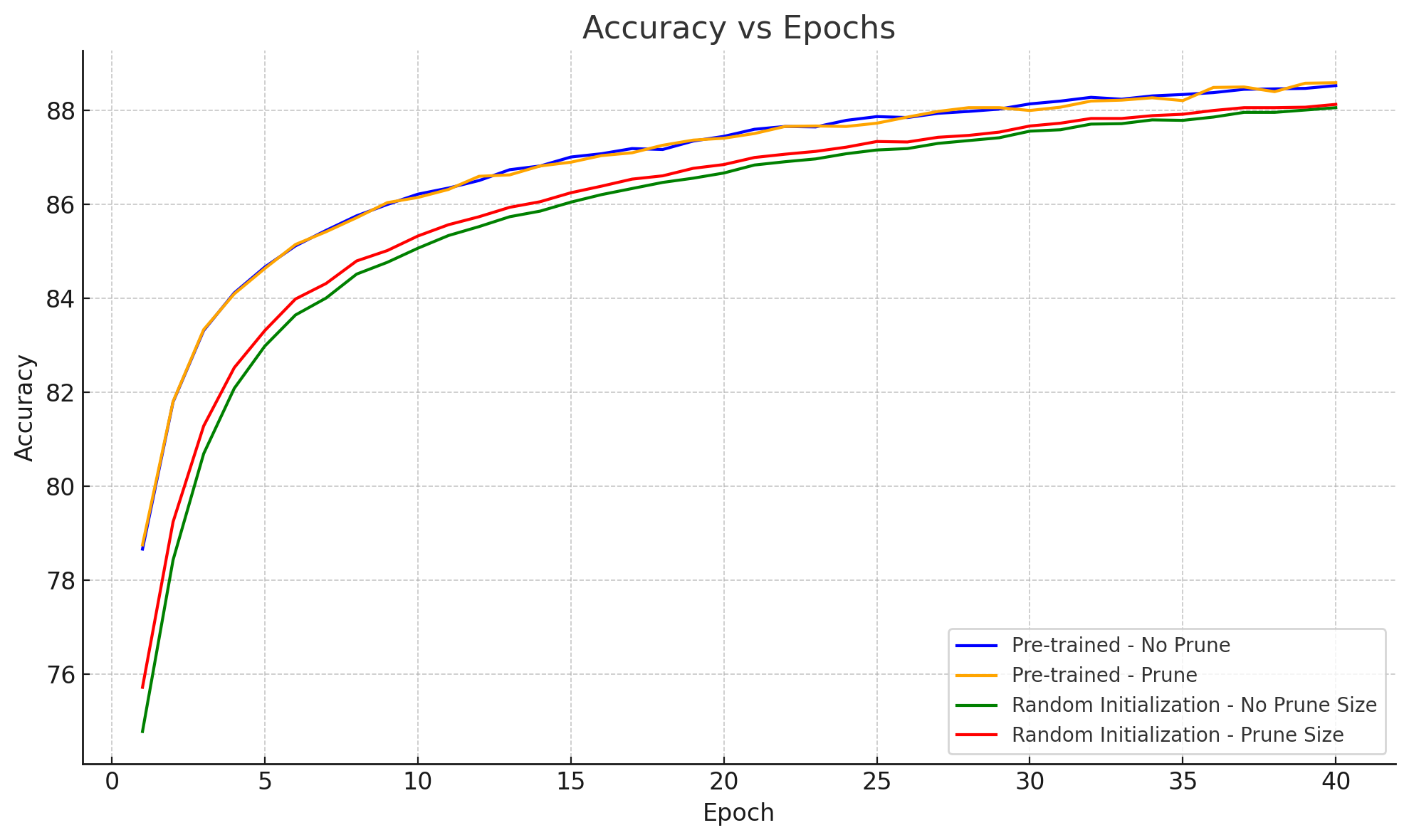}
\caption{Emergence vs Epochs}
\end{minipage}
\hfill
\begin{minipage}{0.48\textwidth}
\centering
\includegraphics[width=\textwidth]{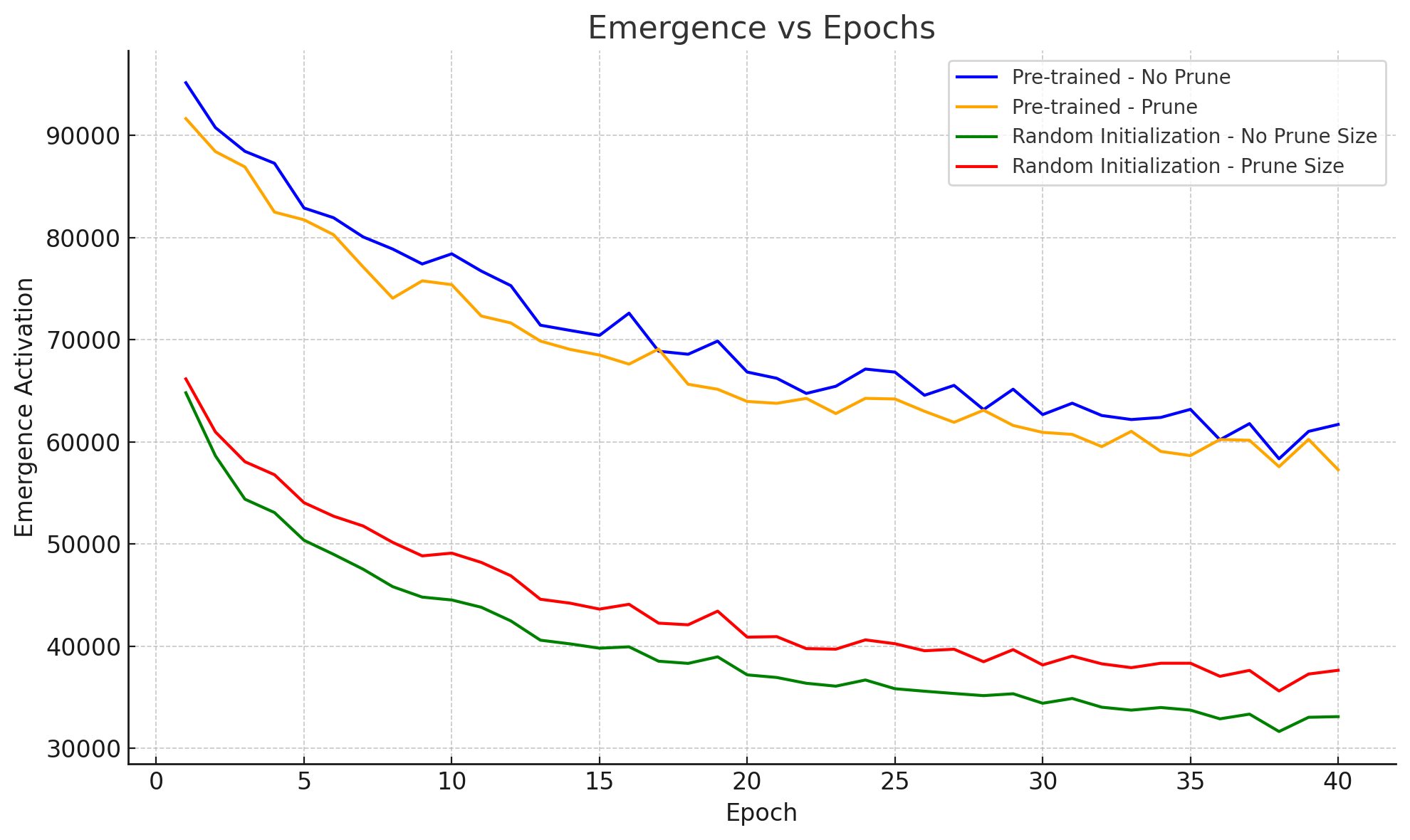}
\caption{Training Accuracy vs Epochs}
\end{minipage}
\end{figure}

The purpose of this experiment was to observe the impact of changing the loss landscape on the emergence of a network. Since  Fashion-MNIST is similar to MNIST, we expected the pruned network to converge faster. This faster convergence indicates higher relative emergence due to less area to explore in the region for local minima. However, since the pruned network had less area to explore, it was more likely that in the long run, the non-pruned network would find a lower minima, aligning with our results.

Although the pruned network converged earlier, it achieved a lower final accuracy in comparison to the non-pruned network, which took longer to converge, yet displayed higher emergence activation and ultimately higher accuracy. This indicates that the potential for emergent traits, facilitated by greater complexity, contributes to the network's ability to achieve higher performance. These results support our idea that pruned networks have lower potential in comparison to non-pruned networks, leading to lower measured values of emergence.

\subsection{CNN Experiments with CIFAR-10 and MNIST}
The experiments conducted on CNNs using CIFAR-10 and MNIST datasets further validate our hypothesis. Similar to the observations in MLPs, we observed a convergence of emergence in CNNs that aligned with the eventual convergence of accuracy. Emergence activation decreased significantly from 634,398 in epoch 1 to zero in the later epochs, indicating that no new emergent traits were forming. This suggests that zero emergence is a reliable indicator that the network will not experience future significant improvements in performance. When emergence is low, accuracy improvements are minimal and appear random, indicating that the network is nearing its full potential. This highlights the importance of emergence as a predictor of the network’s learning capability and its potential for future performance gains.


\subsection{Discussion}
Our study highlights the critical role of emergence in understanding and optimizing neural network performance. Emergence, defined by the connectivity and interactions between active and inactive nodes, serves as a predictor of the network's potential to develop complex, high-performing traits. The results demonstrate that higher levels of emergence correlate with improved trainability and final accuracy, suggesting that networks with greater complexity are better equipped to navigate the loss landscape effectively. Pruning significantly impacts both emergence and performance. By reducing network complexity, pruning decreases absolute emergence, leading to faster convergence. However, the relative emergence—emergence normalized by the network's size—increases in pruned networks, indicating a high density of local minima within their loss landscape. While pruned networks converge more quickly, they achieve lower final accuracy compared to non-pruned networks, which maintain higher levels of absolute emergence and ultimately higher performance. This trade-off underscores the importance of balancing complexity and trainability in neural network design. The predictive capability of emergence is another significant finding. As training progresses, a decrease in emergence correlates with the network's approach to convergence. When emergence drops to zero, further significant improvements in accuracy are unlikely, marking an optimal point to terminate training. This insight can optimize computational resources and streamline training processes. Understanding the role of emergence allows for more informed decisions regarding network complexity and pruning strategies, ultimately contributing to the development of more efficient and effective neural network architectures.

Future research should focus on further validating these findings across different network architectures and datasets. Additionally, exploring the theoretical underpinnings of emergence in more depth, including its mathematical modeling and spatial representation within the loss landscape, will provide a stronger foundation for applying these concepts in practical scenarios. Investigating the impact of various pruning techniques and their timing relative to the training process will also offer valuable insights into optimizing neural network performance.

\section{Conclusion}
In this paper, we have investigated the concept of emergence in artificial neural networks, emphasizing its theoretical foundations and empirical validation. We demonstrated that emergence, defined by the connectivity between active and inactive nodes, serves as a robust predictor of network performance. 
Our experiments
validate that higher emergence correlates with improved trainability and accuracy. We also explored the implications of network complexity and its spatial representation within the loss landscape, revealing that higher emergence indicates a more effective navigation of the loss landscape. Furthermore, we examined the effects of pruning on emergence and network performance, showing that while pruning enhances training efficiency, it typically results in lower final accuracy. Our work, inspired by recent theoretical advancements, provides new insights into the role of emergence in neural network performance, offering significant implications for the design and optimization of efficient neural network architectures.
\bibliography{acml24}
\bibliographystyle{plain}

\end{document}